\pdfoutput=1
\documentclass[11pt]{article}

\usepackage[final]{acl}

\usepackage{times}
\usepackage{amsmath}
\usepackage{amssymb}
\usepackage{amsthm}
\usepackage{latexsym}
\usepackage{booktabs}
\usepackage{multirow}
\usepackage{subcaption}
\usepackage[most]{tcolorbox}

\usepackage{graphicx}
\usepackage{caption}
\usepackage{booktabs}

\usepackage[T1]{fontenc}
\usepackage[utf8]{inputenc}
\usepackage[utf8]{inputenc}
\usepackage[greek,english]{babel}
\usepackage{microtype}

\usepackage{inconsolata}

\usepackage{graphicx}

\title{Assumed Identities: Quantifying Gender Bias in Machine Translation of Gender-Ambiguous Occupational Terms}

\author{Orfeas Menis Mastromichalakis\textsuperscript{1}, Giorgos Filandrianos\textsuperscript{1,2}, \\ \textbf{Maria Symeonaki}\textsuperscript{3} \and \textbf{Giorgos Stamou}\textsuperscript{1} \\
\textsuperscript{1}School of Electrical and Computer Engineering, \\ National Technical University of Athens, Greece \\
 \textsuperscript{2}Instituto de Telecomunicações, Portugal \\
  \textsuperscript{3}Department of Social Policy, Panteion University of Social and Political Sciences, Greece \\
   \texttt{\{\href{mailto:menorf@ails.ece.ntua.gr}{menorf},\href{mailto:geofila@ails.ece.ntua.gr}{geofila}\}@ails.ece.ntua.gr,}
  \texttt{\href{mailto:msymeon@panteion.gr}{msymeon@panteion.gr}, \href{mailto:gstam@cs.ntua.gr}{gstam@cs.ntua.gr}}}
\newtheorem{definition}{Definition}
\begin{document}
\maketitle
\begin{abstract}
Machine Translation (MT) systems frequently encounter gender-ambiguous occupational terms, where they must assign gender without explicit contextual cues. While individual translations in such cases may not be inherently biased, systematic patterns—such as consistently translating certain professions with specific genders—can emerge, reflecting and perpetuating societal stereotypes. This ambiguity challenges traditional instance-level single-answer evaluation approaches, as no single gold standard translation exists. 
To address this, we introduce GRAPE, a probability-based metric designed to evaluate gender bias by analyzing aggregated model responses.  Alongside this, we present GAMBIT, a benchmarking dataset in English with gender-ambiguous occupational terms. Using GRAPE, we evaluate several MT systems and examine whether their gendered translations in Greek and French align with or diverge from societal stereotypes, real-world occupational gender distributions, and normative standards\footnote{Our code is available at \url{https://github.com/ails-lab/assumed-identities}, and the GAMBIT dataset is publicly available at \url{https://huggingface.co/datasets/ailsntua/GAMBIT}.}.
\end{abstract}

\section{Introduction}
\begin{figure}
    \centering    \includegraphics[width=0.9\linewidth]{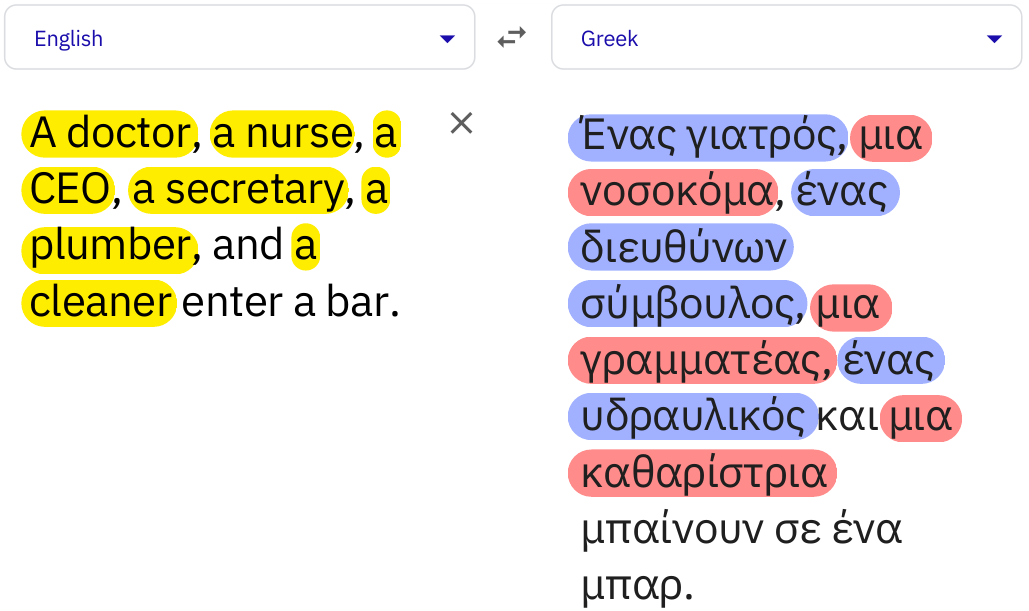}
    \caption{An example of gender stereotypes reflected in a translation from English to Greek (Google Translate). All occupations in the source text (highlighted in yellow) are gender-ambiguous, while target terms highlighted in blue indicate masculine grammatical forms and those in red indicate feminine forms.}
    \label{fig:google-trasnlate-example}
\end{figure}

% MT and occupational gender biases. Examples from Google Translate
Machine Translation systems have become indispensable tools for cross-linguistic communication, yet they frequently exhibit gender biases that reinforce societal stereotypes \cite{blodgett-etal-2020-language, menis2025gender}. In the labour market, where gender disparities persist, such biases are particularly concerning. 
For example, as illustrated in Figure~\ref{fig:google-trasnlate-example},  Google Translate\footnote{\url{https://translate.google.com/}} systematically assigns masculine grammatical forms to occupations traditionally dominated by men or stereotypically perceived as masculine (e.g., CEO, doctor, plumber), and feminine forms to those commonly associated with women (e.g., nurse, secretary, cleaner) when translating gender-ambiguous inputs from English to Greek.
This is not an isolated case, but a consistent pattern across most MT systems \cite{alvarez-melis-jaakkola-2017-causal, escude-font-costa-jussa-2019-equalizing}. 
These biases extend beyond language, subtly validating and reinforcing occupational segregation by shaping perceptions of gender roles. This, in turn, influences hiring practices, career aspirations, and wage disparities, further entrenching systemic inequalities in the workforce \cite{EU_GenderEqualityStrategy}. Addressing these biases is essential to ensure that MT systems contribute to fair representations of professions rather than perpetuating historical and cultural stereotypes.

% Existing approaches to such biases. Introduce the problem (problem statement) pointing out the notable gap in literature 

Evaluating occupational gender bias in MT systems is particularly challenging when the occupational terms are gender-ambiguous. When translating from genderless (e.g., Finnish or Turkish) or notional gendered languages (e.g., English) into languages with grammatical gender (e.g., Greek or French), MT systems often must make assumptions and assign gender, as preserving ambiguity or using gender-neutral language is not always feasible or stylistically appropriate. 
Unlike explicit biases, such as misgendering occupations with clear gender markers in the input, which can be directly flagged as incorrect, these cases exhibit a unique duality: a single translation—e.g., translating ``the actor'' as \emph{``l’acteur''} (masculine) or \emph{``l’actrice'} (feminine) in French—is not biased or unbiased in isolation, as both choices are equally valid in the absence of contextual cues. However, when examined in aggregate, systematic patterns may emerge, revealing a model’s predisposition to associate certain professions with specific genders. This renders traditional instance-level single-answer quality \cite{papineni2002bleu, lin2004rouge} or bias \cite{stanovsky-etal-2019-evaluating} evaluation approaches unsuitable, as they fail to capture these broader distributional trends in the absence of a gold standard.
% Established MT evaluation metrics \cite{papineni2002bleu, lin2004rouge}, including some gender bias evaluation approaches \cite{stanovsky-etal-2019-evaluating} typically rely on explicit correctness criteria, comparing each translation against a single reference gold standard. This approach is ill-equipped to assess biases that only become apparent through repeated associations over a dataset, since such a single gold standard cannot exist. 
% This limitation presents a critical gap in the literature: current approaches lack the mechanisms to detect and quantify biases that manifest at a systemic level rather than in individual instances. 

% Our approach
In this work, we shift the focus from isolated translations to aggregated model behavior, enabling the detection of systematic gender biases that may not be evident at the individual sentence level. We propose a methodology to detect, classify, and quantify gender assignments in the translation of gender-ambiguous occupational terms.
To support this evaluation, we introduce GAMBIT (Gender-AMBIguous occupaTions), a benchmarking dataset of English texts containing occupational terms expressed in a gender-neutral or ambiguous way.
Our approach identifies gender assignments by comparing source texts with their translations and aggregates gendered outputs across multiple instances. Occupations are grouped using the International Standard Classification of Occupations (ISCO-08)\footnote{\url{https://ilostat.ilo.org/methods/concepts-and-definitions/classification-occupation/}}, to enable analysis at varying levels of abstraction and account for lexical variations. ISCO-08 is an internationally recognized system for classifying occupations endorsed by the International Labour Organisation (ILO). It provides a hierarchical structure that categorizes jobs into four levels of increasing granularity, using a digit-based coding system. At the highest level, occupations are grouped into broad categories, which are then divided into more specific subcategories at lower levels. For example, the top-level category ``Professionals'' (code 2) includes subcategories such as ``Science and Engineering Professionals'' (code 21) and ``Health Professionals'' (code 22), that further divide into ``Medical Doctors'' (code 221), ``Nursing and Midwifery Professionals'' (code 222), and others. Each category in ISCO-08 is accompanied by detailed descriptions, examples of occupations, and other relevant information, providing a comprehensive framework for analyzing and comparing jobs across different countries and industries.
GAMBIT spans the entire ISCO-08 taxonomy, ensuring broad occupational coverage.
To quantify bias, we introduce GRAPE, a probability-based metric that measures divergence from reference distributions, such as idealized gender parity or real-world labor statistics. We apply our framework to translations from English into Greek and French, two languages with grammatical gender but from different language families, comparing outputs against both normative standards and empirical labor data. Our approach offers a scalable and interpretable framework to evaluate gender bias in MT, offering insights into how translation systems reflect, reinforce, or potentially challenge societal patterns of occupational gender representation when facing ambiguity.

\section{Related Work}

Research on gender bias in NLP has explored a broad range of tasks and provided valuable insights \cite{bolukbasi2016man, lu2020gender}, but our focus is on Machine Translation \cite{savoldi-etal-2021-gender, vanmassenhove20249}, where gender bias remains a pressing issue with significant societal impact \cite{savoldi2024harm}. Numerous case studies have highlighted the prevalence and consequences of gender bias in MT across languages and cultural contexts \cite{ rescigno2020case,  farkas2022measure, ghosh2023chatgpt, paolucci-etal-2023-gender, kostikova-etal-2023-adaptive, piazzolla2023good}, emphasizing the need for effective evaluation and mitigation. Moreover, critiques of existing quality metrics reveal that traditional evaluation methods often fail to capture gender disparities adequately \cite{zaranis2024watching}.
To tackle this, researchers have developed resources and methods that target gender bias in MT. This includes Knowledge Graphs that offer structured, contextual information for bias analysis \cite{mastromichalakis2024gost}, multilingual benchmarks \cite{currey-etal-2022-mt}, and studies on language-specific challenges such as gender-neutral pronouns \cite{cho-etal-2019-measuring}. Alongside, mitigation efforts \cite{sun-etal-2019-mitigating} explore model fine-tuning, data balancing, and adaptive learning \cite{escude-font-costa-jussa-2019-equalizing, saunders-byrne-2020-reducing,  costa-jussa-de-jorge-2020-fine}, with recent work also focusing on gender-neutral and gender inclusive translation strategies \cite{piergentili-etal-2023-gender, lardelli-gromann-2023-gender} and benchmarking such approaches \cite{piergentili-etal-2023-hi, lardelli-etal-2024-building, gkovedarou2025gender}.

Our work studies occupational gender bias in translating gender-ambiguous inputs, adding to ongoing research on gender bias in NLP with a focus on occupations and the labor market \cite{tal-etal-2022-fewer, gorti2024unboxing}. Ambiguity has also been studied in other NLP tasks, such as Question Answering \cite{li-etal-2020-unqovering, parrish-etal-2022-bbq} and coreference resolution \cite{rudinger-etal-2018-gender, zhao-etal-2018-gender}, where multiple plausible interpretations reveal the influence of stereotypes. For example, \citet{kotek2023gender} examine ambiguous coreference inputs in LLMs, where no single ground truth exists. Their aggregated analysis of role, trait, and occupation associations exposes stereotypical patterns, aligning with our approach of studying model behavior at an aggregated level to detect subtle biases.
In MT, one way to handle ambiguity is by generating all grammatically correct gendered translations \cite{garg2024generating}, a strategy used by some commercial systems. While inclusive, this approach is limited to setups that allow multiple outputs and faces scalability challenges as multiple ambiguities exponentially increase possible translations. Other approaches disambiguate inputs before translation \cite{vanmassenhove-etal-2018-getting}, which however requires some structural or semantic hints that allow the disambiguation of gender. This is the case for some challenge sets like WinoMT \cite{stanovsky-etal-2019-evaluating} where gender can be disambiguated via correference, or the MuST-SHE corpus \cite{bentivogli-etal-2020-gender} that includes audios and transcripts, where the inputs have a correct gender resolution due to gender cues that are recoverable from audio (e.g., speaker’s voice) or textual context (e.g., pronouns, named entities).

In contrast, our study focuses on ambiguous cases without disambiguating cues, allowing inherent stereotypical associations and biases to emerge naturally. Our inputs are deliberately designed to have multiple plausible interpretations without a single correct answer. 
Gender ambiguity in MT has been explored through a range of challenge sets and benchmarks, yet most existing efforts remain limited in scope, scale, or evaluation depth.
gENder-IT \cite{vanmassenhove-monti-2021-gender} introduced a manually curated English–Italian challenge set stemming from MuST-SHE, covering natural gender phenomena, including occupation-related examples. While it includes truly ambiguous instances, the dataset remains limited in scale (694 sentences), treats each sentence in isolation, and lacks a structured evaluation methodology. A follow-up study \cite{vanmassenhove20249} used a subset of gENder-IT to assess ChatGPT's performance, but provided only a brief and high-level analysis.
Concurrent to our work, \citet{hackenbuchner2025genderous} introduced GENDEROUS, a handcrafted dataset of sentences with statistically stereotypical occupational nouns and gender-inflected adjectives, and examined the effect of gender-ambiguous inputs.
Other works have explored bias through contrastive sentence pairs. \citet{gonen-webster-2020-automatically} for instance, generate minimal pairs differing by a single human-related noun to expose gender asymmetries in translations.
In a different approach, \citet{prates2020assessing} examine gender bias in Google Translate using simple templates. While their aggregated analysis shares similarities with ours, the reliance on templated inputs and the focus on a single MT system restricts the generalizability of their findings.
Our work on the other hand, introduces a comprehensive evaluation framework that goes beyond challenge sets. It covers a broad range of occupations based on ISCO classifications, provides rich contextual texts rather than isolated sentences, enables structured, multilingual evaluation through interpretable, statistics-informed metrics, and evaluates a variety of MT systems.

\section{Detecting Gender Assignments}
\label{sec:method}

Our approach focuses on detecting gender assignments in the translation of occupational terms when the source gender is ambiguous. By gender assignment, we refer to cases where the translation introduces a masculine or feminine form not specified in the source. Given a gender-ambiguous input, we feed it to an MT system and analyze the output to determine the gender of any translated occupational terms. This process follows an approach inspired by the ``LLM-as-a-judge'' paradigm \cite{li2025generationjudgmentopportunitieschallenges, gu2025surveyllmasajudge}, a framework that has previously been applied to evaluate gender bias in machine translation systems and textual corpora~\cite{derner2024leveraging, piergentili2025llm}, and consists of two main steps: (1) detecting occupations in the translation, and (2) identifying their gender. 

For the first step, we employ an LLM-based component in a few-shot setup, prompting it to extract all explicitly mentioned occupation titles in the text. During development, we identified two main types of hallucinations and designed targeted strategies to mitigate them.
The first type occurs when the LLM \textit{identifies occupations that are not present in the text}. For example, in the sentence: ``The supplier complained to the call center,'' the LLM may incorrectly detect the occupation ``Customer Service Representative'', even though it is not explicitly mentioned. Note that this example is given in English for illustrative purposes; in practice, such cases only arose in the French and Greek translations, since occupation and gender detection were not performed on the English source texts (which were already provided by the GAMBIT dataset). To address this, we instructed the LLM to provide both the detected occupation titles and their corresponding in-text occurrences. We then applied fuzzy string matching to verify whether the detected terms appeared in the text. If the similarity fell below a predefined threshold, the term was discarded as a hallucination.
The second type of hallucination involves the LLM \emph{incorrectly identifying non-occupational terms as occupations}. This issue was particularly prevalent in cases where no occupations were present in the input text.
For instance, in ``She is a master in her craft.'', the word ``Master'' was wrongly detected as an occupation. To address this, we modified the prompt to require the LLM to also generate a short description for each detected occupation. This description serves as a verification step to check whether the detected occupation matches any ISCO-08 entry. 
To do this comparison, we used an embedding-based approach. We converted both the LLM-generated descriptions and the ISCO-08 descriptions into embeddings and applied cosine similarity to find the closest match. Following the method from \citet{li-li-2024-aoe}, we used angle-based embeddings to map the descriptions into a common latent space. If the similarity score between the LLM’s description and any ISCO-08 occupation was below a threshold, the detected occupation was discarded. This step improved both accuracy and consistency by ensuring alignment with ISCO-08 occupations. In our case, the comparison was limited to a known set of candidate occupations from the source text, which simplified the task and made thresholding more efficient.

The second step of our approach involves identifying the gender of the detected occupations, assigning to them one of the three labels: \textit{``Masculine,'' ``Feminine,''} or \textit{``Not Clear''}. This is done using the same LLM, within the same session. After detecting an occupation, the LLM is prompted to assign a gender label to each identified occupation.
 
This pipeline allows us to detect and measure gender assignments between source and translated texts. These assignments are then aggregated using the evaluation framework described in Section~\ref{sec:framework} to quantify the model’s gender bias. Technical  implementation details are provided in Appendix~\ref{app:tech-details}, and all prompts are listed in Appendix~\ref{app:prompts}.

% [Probably the following fits better in the next section (?)]
% This evaluation approach seems to have greater applicability for translations from genderless languages to notional gender and grammatical gender languages, as well as translations from notional gender to languages with grammatical gender (so from a language with "less" gender to a language with "more" gender if you could put these 3 types of languages in a scale of "tenderness"). This is because n these cases there are more cases of ambiguous inputs (gender not clear) that would need to be translated to a gendered form. There are also few examples from grammatical gender languages to notional gender languages, but they are very specific cases. However, our approach would still be valid. 

\section{Evaluation Framework}
\label{sec:framework}
In this section, we present an evaluation framework to study occupational gender bias in Machine Translation systems when handling gender-ambiguous inputs.
% Traditional evaluation methods often rely on single-answer correctness, which is inadequate for ambiguous scenarios where multiple gendered translations could be equally valid. Although some existing approaches investigate multiple outputs, the single-output setup remains the most common in practice, therefore it is crucial to examine systems’ behavior in this context. 
Our goal is to quantify gender bias by analyzing the distribution of gendered translations across these ambiguous cases, revealing patterns of bias that instance-level evaluations may overlook. 
% When no instance-level correct or ground truth translations exist, a fundamental question arises: what is the desirable behavior of a system? In ambiguous cases, traditional accuracy metrics fall short, as they presume the existence of a single correct answer. Systems could address ambiguity by producing multiple outputs, each representing a grammatically correct gender alternative, when the task and setup allow it. Alternatively, systems could indicate uncertainty or request clarification—a strategy more suitable for interactive applications like chatbots, where user feedback can resolve ambiguity. Another option is for the system to abstain from producing a gendered translation in ambiguous cases. However, this prioritizes harmlessness over helpfulness, potentially limiting the system's usability and utility. A further possibility is for the system to maintain ambiguity in the output without making gendered assumptions, but this is not always feasible due to the grammatical and morphological constraints of some languages.

In real-world applications, MT systems typically produce a single output, forcing a choice when ambiguity is present. In these cases, the system makes an implicit assumption, raising the question of how this decision should be evaluated. Here, two, sometimes competing, perspectives emerge: \emph{normative correctness} and \emph{predictive accuracy} \cite{deery2022bias}. Normative correctness evaluates system behavior against an idealized standard of fairness, such as gender parity. Predictive accuracy, on the other hand, assesses how well the system reflects a reference distribution, such as real-world gender statistics for a given occupation.
% By adopting both perspectives, we aim to provide a more nuanced and comprehensive evaluation of gender bias in MT systems, balancing ethical considerations with empirical realities.

Since our approach aggregates behavior across multiple outputs and we do not expect consistent behavior across all occupations, it is essential to group outputs that refer to the same occupation(s). This will allow a more detailed investigation, identifying the model's associations between specific occupations and gender. However, simple keyword-based clustering is inadequate due to variation in how occupations are expressed, and  could lead to fragmented or inconsistent clusters. Moreover, many occupations are semantically related, while others differ significantly. Analyzing each occupation in isolation limits the ability to draw generalizable conclusions. 

To address this, we adopt ISCO-08 as our occupation taxonomy. It allows us to cluster, organize, and analyze translations at multiple levels of abstraction, beginning with detailed (4-digit) categories. This enables us to investigate gender bias both at the level of specific occupations and across broader occupational groupings. Beyond clustering, ISCO-08 also captures hierarchical relationships among occupations, which we use to structure our analysis and identify patterns of bias that span related roles.
Using ISCO-08 ensures consistency and comparability across occupations, supports structured and meaningful generalization, and aligns our framework with international standards. This, in turn, facilitates comparisons with real-world labor statistics and improves the interpretability and applicability of our findings.

\subsection{Metrics}
To quantify gender bias in MT and compare model outputs against reference distributions, we introduce the \textit{Gender RAtio ProbabilitiEs} (GRAPE). This metric measures how the likelihood of generating masculine or feminine forms for gender-ambiguous terms diverges from a chosen reference distribution. 
% Since we are comparing two binary outcomes (masculine vs. feminine), these probabilities are interpreted as Bernoulli probabilities.

% In order to quantify gender bias of an MT system, and be able to compare the system's outputs against normative standards or reference distributions, following the normative correctness and predictive accuracy perspectives discussed above, we introduce GPD (and nGPD), a probability-based metric (and its normalized form). 
% In this subsection, we propose metrics to evaluate gender bias in MT systems for occupational terms in ambiguous scenarios, where the model has to assign gender to the occupational term.

% In such cases, the ideal behavior would be to maintain this ambiguity in the translation. However, when translating from genderless or notional gender languages to grammatical gender languages, maintaining this ambiguity is often not feasible due to grammatical and morphological constraints. Consequently, the MT system must assign a gender to the occupational term, revealing potential biases in gender assumptions. Our metric is designed to quantify this bias. 
% The proposed metric compares the observed probability of generating masculine (or feminine) forms in gender-ambiguous inputs against a reference probability. Since we are comparing two binary outcomes (masculine vs. feminine), the probabilities can be interpreted as Bernoulli probabilities.

\begin{definition}[Gender RAtio ProbabilitiEs (GRAPE)]
Let \( M \) be a set of source–target text pairs, where each source contains a gender-ambiguous term. Let \( p_{\text{m}} \) be the observed probability of generating a masculine form in \( M \), and \( p_{\text{f}} = 1 - p_{\text{m}} \) the probability of generating a feminine form. Let \( p^{\text{ref}}_{\text{m}} \) denote the reference probability for the masculine form, and \( p^{\text{ref}}_{\text{f}} = 1 - p^{\text{ref}}_{\text{m}} \) for the feminine.

% Let \( M \) be a set of text pairs, where each consists of a source text containing a gender-ambiguous occupational term and its corresponding translation. We define \( p_{\text{masc}} \) as the observed probability of generating a masculine form in \( M \), and \( p_{\text{fem}} = 1 - p_{\text{masc}} \) as the probability of generating a feminine form. Let \( p^{\text{ref}}_{\text{masc}} \) be the reference probability for the masculine form, and \( p^{\text{ref}}_{\text{fem}} = 1 - p^{\text{ref}}_{\text{masc}} \) for the feminine.
\noindent
\textit{GRAPE} is defined as:
% \begin{equation}
% \text{GPD}^{\text{ref}}_{g}(M) = p_{g} - p^{\text{ref}}_{g}, g\in \{\text{masc}, \text{fem}\} 
% \end{equation}

% \noindent And its normalized form, \textit{nGPD}, is: 
\begin{equation}
\text{GRAPE}^{\text{ref}}_{g}(M) = \frac{p_{g} - p^{\text{ref}}_{g}}{p^{\text{ref}}_{g}}, g\in \{\text{m}, \text{f}\} 
\end{equation}

\noindent Positive values indicate bias toward the respective gender, while negative values against it.

% \noindent Positive values indicate a bias towards the respective gender form, while negative values indicate a bias against it. In this work we use as reference distribution either the equal distribution between the two genders (\emph{ref} = parity), or real-world statistics (\emph{ref} = real)
\end{definition}

\noindent
Intuitively, GRAPE measures the relative difference between the model’s output probability for a gendered form and the corresponding reference probability. For example, $\text{GRAPE}^{\text{ref}}_{\text{m}}(M) = 1.0$ implies that the system generates masculine forms twice (100\%) more often than expected based on the reference. These metrics quantify both the \textit{direction} and \textit{magnitude} of gender bias.

% Intuitively, GPD measures the difference between the probability that the model will output a masculine (or feminine) form and the probability of an individual,instance being male/masculine (or female/feminine) in the reference distribution. Its normalized version, nGPD,  indicates the relative deviation as a percentage, showing how much the model’s behavior diverges from the reference distribution in proportion to the reference probability. The metrics provide insights into the direction and magnitude of gender bias in MT outputs. For example, $\text{nGPD}_{masc}(M) = 0.2$ implies that the MT system is 20\% more likely to produce a masculine form compared to the reference distribution. 
% Conversely, a negative value would indicate a bias towards the feminine form.

% The metric is calculated over a set of input-output pairs, $M$. For the results to have semantic and research value, the set should have a meaningful grouping, such as all ambiguous inputs related to a particular occupation. In our approach, we cluster the input-output pairs based on ISCO (International Standard Classification of Occupations) groups, allowing us to analyze gender bias within specific occupational categories.

Although MT outputs are not always strictly binary in gender, maintaining ambiguity is often impractical in gendered languages due to grammatical and morphological constraints. Some languages, such as Greek, include epicene occupational terms that are identical for masculine and feminine forms. However, even in these cases, gendered pronouns and articles often reveal gender. Additionally, historically masculine epicene terms (e.g., \textgreek{βουλευτής}) are increasingly complemented by feminine forms (e.g., \textgreek{βουλεύτρια}), reflecting evolving usage in discourse and media. These factors make it difficult for MT systems to preserve gender ambiguity in translation. While some systems use gender-neutral strategies (e.g., they/them in English), such approaches are not yet widespread or standardized. In our evaluation (Section~\ref{sec:results}), gender ambiguity was preserved in fewer than 15\% of instances on average. Furthermore, reference distributions (e.g., parity or real-world statistics) typically lack a neutral category, making it difficult to include gender-neutral outputs in our framework. We therefore focus on binary gender forms and leave the integration of neutrality to future work.

% It's important to note that a model's output is not necessarily binary in terms of gender. It is possible to maintain the ambiguity even in languages with grammatical gender by either avoiding gendered terms and pronouns, or by using gender-neutral language, like the use of the pronouns they/them in English, to refer to persons of unknown or non-binary gender. Although such behavior could be considered desirable in some setups, maintaining ambiguity without the use of gender-neutral language is often not feasible due to grammatical and morphological constraints of the language, and the use of gender-neutral language is not yet widespread or standardized across all languages. While recent research, as noted in our related work section, has begun to explore gender-neutral translation strategies, most current MT systems do not actively incorporate such approaches. This is also evident in our results in Section~\ref{sec:results} where the models evaluated maintained gender ambiguity (i.e. the occupations remained gender-ambiguous or gender-neutral) in less than 10\% of the examples in both languages, so this does not significantly affect the results of our evaluation. Although neutral gender could be incorporated to our metric, doing so would prevent us from comparing model outputs to reference distributions as in most cases, reference distributions (like parity or real-world statistics) do not have a neutral gender option. Therefore, such cases are not considered in this framework, but we leave as future work the incorporation of gender-neutrality in this framework.   

The choice of reference distribution is central to interpreting the metrics, and we adopt two perspectives:
\begin{itemize}
  \item \textbf{Normative Correctness}: Assumes ideal gender parity by setting \( p^{\text{ref}}_{\text{m}} = p^{\text{ref}}_{\text{f}} = 0.5 \). This baseline reflects an expectation of equal representation. In this case we use \emph{ref}=parity. 
  \item \textbf{Predictive Accuracy}: Uses empirical data (e.g., labor statistics) to reflect actual gender distributions across occupations. This enables contextual evaluation grounded in real-world demographics. In this case we use \emph{ref}=real. 
\end{itemize}

\noindent
By applying both perspectives, our evaluation framework captures different dimensions of fairness: one based on equality, the other on realistic alignment.

\subsection{Benchmarking Dataset}
\label{sec:benchmarking}
To enable a comprehensive evaluation of MT systems across the full range of occupations in the ISCO taxonomy, we created GAMBIT, a benchmarking dataset containing English texts with gender-ambiguous occupational terms. Existing datasets \cite{rudinger-etal-2018-gender, zhao-etal-2018-gender, stanovsky-etal-2019-evaluating} lacked sufficient occupational coverage, particularly in gender-ambiguous contexts, so we opted to generate the dataset using large language models (LLMs), followed by thorough manual review for quality assurance. All generated instances were validated by domain experts, namely PhD holders in gender studies, social policy, and sociology, with established expertise in occupation-related topics. Validation involved discarding any texts that were not fluent or where occupational terms were not gender-ambiguous. The experts carried out this task as part of their work in a funded project and were compensated according to national standards. Before generating GAMBIT, we attempted to build a dataset from real-world data by processing over 250,000 random texts from the WMT\footnote{\url{https://huggingface.co/datasets/wmt/wmt14}} and C4\footnote{\url{https://huggingface.co/datasets/allenai/c4}} \cite{dodge-etal-2021-documenting} datasets. However, this approach yielded data for only 43 ISCO unit groups (less than 10\% of the 436 total), with limited textual diversity and repetitive patterns that could introduce bias. This made artificial generation the only viable approach for ensuring both full occupational coverage and a variety of textual styles.

For the generation, we used Claude 3.5 Sonnet\footnote{Model ID: anthropic.claude-3-5-sonnet-20241022-v2:0}. Detailed information about the prompts is provided in Appendix~\ref{app:prompts}. We collected all occupational titles from each 4-digit ISCO-08 class and generated multiple examples per occupation, varying by text format.
GAMBIT consists of 9,805 English samples, averaging 22.5 texts per occupation, distributed evenly across five formats: short stories, brief news reports, short statements, short conversations, and short presentations (1,961 samples per format). Detailed statistics on character and word length are provided in Appendix~\ref{app:gambit}.
The dataset is designed to support gender bias evaluation for any language pair with English as the source language. Adapting the methodology proposed in Section~\ref{sec:method} to a different target language requires only minimal changes, as the core detection components rely on LLMs, which are available for most languages.

\subsection{Real World Statistics}
To calculate the reference distribution in the predictive approach of our metrics, we collected real-world labor statistics. Specifically, we collected the gender-based occupational distributions for France and Greece, since our translation tasks involve English to French and English to Greek, respectively. Although both languages are also spoken in other parts of the world, we focused on these countries as representative examples to demonstrate how our approach can incorporate real-world demographics.
 
We analyzed raw microdata drawn from the European Union Labour Force Survey (EU-LFS)\footnote{\url{https://ec.europa.eu/eurostat/web/microdata/european-union-labour-force-survey}}. This large-scale sample survey provides quarterly and annual statistics on labor participation and inactivity among individuals aged 15 and older,  using standardized definitions and the ISCO-08 classification to ensure cross-country comparability. In particular, we calculated the gendered occupational distributions at the ISCO-08 3-digit level for both countries over the period 2011-2023. This allows us to benchmark MT systems against real-world occupational gender distributions, providing a meaningful reference point for evaluating gender bias.  As these statistics follow the ISCO-08 classification, they align directly with our benchmark dataset, enabling straightforward mapping between the two.

\section{Experiments and Results}
\label{sec:results}
\subsection{Pipeline's Performance}
% First, the performance of the pipeline was tested using traditional datasets in the field, particularly in English, such as Winobias \cite{zhao-etal-2018-gender}. However, due to the restricted set of occupation titles and the specific format of the texts in these datasets, the overall accuracy of the pipeline for both occupation and gender identification was 100\%. 
To evaluate the performance of the pipeline introduced in Section~\ref{sec:method}, we constructed a separate validation dataset in French and Greek.
Existing datasets were either limited to English or covered only a narrow range of occupations. Therefore, we followed the same construction approach as for GAMBIT (see~\ref{sec:benchmarking}), ensuring broad occupational coverage across the ISCO classification and textual variety. The datasets included masculine, feminine, and, where possible, gender-neutral forms of occupations, allowing us to assess how well the pipeline handles gender-specific and ambiguous cases. Each language dataset contains 29,415 texts, with occupations evenly distributed across the ISCO taxonomy. A random sample of approximately 20\% of the data was manually reviewed, and no issues with the text or labels were found. Further details are provided in Appendix~\ref{app:evaluation}.

Table~\ref{tab:performance} presents the pipeline’s performance, reporting accuracy in identifying occupations, detecting gender, and combining both, using Claude 3.5 Sonnet. The results show that the pipeline reliably extracts both occupation and gender information, making it a suitable tool for analyzing gender-related behavior in machine translation systems.

\begin{table}[]
\small
\centering
\begin{tabular}{l|cc|c}
\toprule
   Lang     & Occ. Acc. & Gender Acc. & Overall \\ \midrule
% English & 99.86 &  99.34 &  99.21    \\
French  & 99.93 &   98.30 &  98.30 \\
Greek   & 99.92 & 99.53   &  99.47 \\ \bottomrule
\end{tabular}
\caption{Pipeline accuracy}
\label{tab:performance}
\end{table}

% \begin{table}[]
% \small
% \centering
% \begin{tabular}{l|cc|c}
% \toprule
%    Lang     & Occ. Acc. & Gender Acc. & Overall \\ \midrule
% % English & 96.9 &  95.9 &  92.9    \\
% French  &  90.32 &  96.43 &   87.10 \\
% Greek   & 92.42 & 97.31   &  91.17 \\ \bottomrule
% \end{tabular}
% \caption{Pipeline accuracy on real-world data}
% \label{tab:performance_real}
% \end{table}

% \begin{table}[]
% \small
% \centering
% \begin{tabular}{c|ccc|ccc}
% \toprule
% \multirow{2}{*}{MT} & \multicolumn{3}{c|}{French} & \multicolumn{3}{c}{Greek} \\ \cline{2-7} 
%                     & Masc.     & Fem.    & NC    & Masc.    & Fem.    & NC   \\ \hline

% m2m100 & \textbf{83.31} & 3.37 & 13.31 & \textbf{92.70} & 1.37 & 5.92 \\
% nllb &  \textbf{84.83} & 3.71 & 11.47 & \textbf{91.29} & 4.78 & 3.92  \\
% GT & \textbf{83.36} & 3.37 & 13.27 & \textbf{92.71} & 1.37 & 5.92 \\
% Claude & \textbf{78.82} & 8.02 & 13.16 & \textbf{93.59} & 4.83 & 1.58 \\
% \bottomrule
% \end{tabular}
% \caption{The distribution of male, female, and unclear gender categories across the different MT systems used in the study.} 
% \label{tab:mt_performance}
% \end{table}

\subsection{Analysis of MT systems}
We evaluated several widely used MT systems, including Google Translate, M2M100 \cite{10.5555/3546258.3546365}, and NLLB \cite{costa2022no} with 600 million and 1.2 billion parameters, as well as LLMs like Claude-3.5, and EuroLLM \cite{martins2024eurollm}, prompted to perform translations. Further details on the models and implementation are provided in Appendix~\ref{app:tech-details} and the prompts used for the LLM translations in Appendix~\ref{app:prompts}.

\begin{table*}[]
\small
\centering
\begin{tabular}{l|cccc|cccc}
\toprule
\multirow{3}{*}{MT} & \multicolumn{4}{c|}{\emph{ref}=parity} & \multicolumn{4}{c}{\emph{ref}=real} \\ \cline{2-9} 
& \multicolumn{2}{c|}{French} & \multicolumn{2}{c|}{Greek} & \multicolumn{2}{c|}{French} & \multicolumn{2}{c}{Greek} \\ \cline{2-9} 
                    & m     & f     & m    & f   & m     & f     & m    & f  \\ \hline
NLLB-600M & 0.92 & -0.92 & 0.91 & -0.91 & 0.88 & -0.91 & 0.67 & -0.90 \\
NLLB-1.3B & \underline{0.66} & \underline{-0.66} & \underline{0.58} & \underline{-0.58} & \underline{0.63} & \underline{-0.65} & \underline{0.38} & \underline{-0.51}
 \\
M2M100 & \textbf{0.94} & \textbf{-0.94} & 0.95 & -0.95  & \textbf{0.90} & \textbf{-0.94} & 0.71 & -0.95 \\
EuroLLM-1.7B & 0.86 & -0.86 & 0.78 & -0.78 & 0.82 & -0.85 & 0.56 & -0.75 \\
GT & 0.92 & -0.92 & \textbf{0.97} & \textbf{-0.97} &  0.89 & -0.92 & \textbf{0.72} & \textbf{-0.97}\\
Claude & 0.87 & -0.87 & 0.92 & -0.92  & 0.83 & -0.86 & 0.67 & -0.90\\
\bottomrule
\end{tabular}
% \caption{nGPD with M = GAMBIT. Highest absolute values are depicted in \textbf{bold}, while the lowest are \underline{underlined}.} 
\caption{GRAPE calculated on the whole GAMBIT dataset for the two genders across the different MT systems used in the study. Highest absolute values are depicted in \textbf{bold}, while the lowest are \underline{underlined}.} 
\label{tab:mt_performance}
\end{table*}

\subsubsection{Overall Behavior}
We first examined the overall behavior of the MT systems, focusing on how biased their outputs are, and how they align with gender parity and real-world occupational distributions. Table~\ref{tab:mt_performance} presents GRAPE for masculine and feminine translations across systems, using both ideal parity and real-world data as reference points.
The results show a clear and consistent trend: MT systems overwhelmingly translate gender-ambiguous texts into masculine forms. This confirms previous findings that MT models often adopt a \emph{masculine as default} strategy when gender is unclear \cite{schiebinger2014scientific, vanmassenhove-etal-2018-getting, monti2020gender}.  Notably, this tendency is not unique to automated systems. It reflects broader patterns in human language use, where masculine forms are commonly used in situations of gender ambiguity, not only in translation but also in everyday communication \cite{silveira1980generic, stahlberg2011representation}. This is likely reflected in the training data used for MT systems and LLMs, leading to this bias towards masculine forms. 

A tendency toward extreme gendering is also evident at the level of individual occupations, however not always towards masculine forms. On average, in 374 out of 436 ISCO occupations (4-digit level), the systems assigned one gender in more than 80\% of the translated texts, with the percentage being 100\% (i.e. $GRAPE^{parity}_{m}=1$ or $GRAPE^{parity}_{f}=1$) in 293 of them. Only about 30 occupations showed a more balanced output, with gender assignments falling between 30–70\% (see Appendix~\ref{app:extreme} for per-model breakdowns). While this confirms the dominant masculine bias, as the vast majority of the extreme cases were masculine-dominated, it also points to a broader issue:
% models often settle on a single gender per occupation, regardless of variation in the input texts. 
% In fact, an average of 290 occupations per language and system are consistently translated into masculine forms ($nGPD^{parity}_{masc}=1$), with some systems reaching up to 399. In contrast, only about 3 occupations on average are entirely translated into feminine forms ($nGPD^{parity}_{fem}=1$).
% This behavior suggests that 
models tend to rigidly associate specific occupations with specific genders. In most cases, variation in format, type, and context of the input texts had little effect on the gendered output. This may reflect an underlying tendency of current MT systems to reinforce strong associations learned during training—especially when the task permits or encourages confident, consistent outputs. While such determinism can be useful in many settings, it may also limit the model’s ability to reflect ambiguity or diversity.

% First, we assessed the models’ overall bias against normative standards of equal gender representation as well as real-world statistics.
% Table~\ref{tab:mt_performance} presents the overall nGPD for masculine and feminine translations across the different MT systems both calculated with parity and real data as a reference. The results reveal a consistent pattern, with a dominant tendency to translate gender-ambiguous texts into masculine forms. This observation is aligned to existing works that claim that MT systems have \emph{masculine as default}  \cite{schiebinger2014scientific, vanmassenhove-etal-2018-getting, monti2020gender}, which means that when the model is faced with gender-ambiguous terms, it defaults to masculine form translations. 

% In our analysis, we exclude cases where ambiguity is preserved in the translated text, as this reflects the desired behavior and does not contribute to bias quantification. In all models, the percentage of texts maintaining this ambiguity was relatively low, around 10-15\% for French and less than 10\% for Greek.
\subsubsection{Influence of Gender Stereotypes}
Despite the overall masculine skew, this tendency is not uniform across all occupations. In fact, all models consistently translate a small number of occupations predominantly into feminine forms. 
The occupations translated into feminine forms are largely consistent across all MT systems and both languages. These include stereotypically feminine roles such as `Midwifery Professionals' (2222) and associate professionals (3222), `Nursing Professionals' (2221), and `Cleaning and Housekeeping Supervisors in Offices, Hotels, and Other Establishments' (5151). In contrast, the occupations translated into masculine forms include not only stereotypically masculine roles—such as miners (8111), house builders (7111), and judges (2612)—but also those perceived as gender-neutral, like visual artists (2651) and high school teachers (2330).

This suggests that MT systems tend to translate stereotypically feminine occupations into feminine forms, while defaulting to masculine for both stereotypically masculine and neutral roles. To validate this, we analyzed $GRAPE^{parity}_{m}$ across occupations categorized by gender stereotypes as masculine, feminine, and neutral. While real-world gender distributions are often used as proxies for stereotypes, they are not entirely aligned. Research shows that occupational gender stereotypes may reflect outdated perceptions rather than current workforce statistics, with notable mismatches in certain roles \cite{gygax2016true}. To assess stereotypical perceptions directly, we used ratings from \citet{shinar1975sexual}, who provide stereotype scores (1 to 7) for 129 occupations. Appendix~\ref{app:stereotypes} details how we processed this data to group occupations by perceived gender. Using these groupings, we calculated $GRAPE^{parity}_{m}$ for all models in both languages. The results, shown in Table~\ref{tab:stereotype}, confirm our observations: all systems predominantly use masculine forms for stereotypically masculine and neutral occupations, while showing more balanced or feminine-leaning translations for stereotypically feminine occupations.

\begin{table*}[]
\small
\centering
\begin{tabular}{l|cc|cc|cc}
\toprule
\multirow{2}{*}{MT} & \multicolumn{2}{c|}{masculine} & \multicolumn{2}{c|}{neutral} & \multicolumn{2}{c}{feminine} \\ \cline{2-7} 
& French & Greek & French & Greek & French & Greek \\ \hline
NLLB-600M & 0.96 & 0.94 & 0.94 & 0.94 & 0.07 & 0.14  \\
NLLB-1.3B & 0.79 & 0.75 & 0.67 & 0.72 & -0.19 & 0.14  \\
M2M100 & 0.95 & 0.97 & 0.97 & 0.95 & 0.42 & 0.34  \\
EuroLLM-1.7B & 0.95 & 0.92 & 0.87 & 0.78 & -0.09 & -0.22 \\ 
GT & 0.96 & 0.99 & 0.95 & 0.99 & 0.09 & 0.46  \\
Claude & 0.95 & 0.97 & 0.86 & 0.95 & -0.23 & -0.04 \\  
% NLLB-1.3B & 0.79 & 0.75 & 0.67 & 0.72 & -0.19 & 0.14 \\ 
\bottomrule
\end{tabular}
\caption{$GRAPE^{parity}_{m}$ for stereotypically masculine, neutral, and feminine occupations. } 
\label{tab:stereotype}
\end{table*}

% \begin{table}[]
% \small
% \centering
% \begin{tabular}{c|cc}
% \toprule
% MT & French & Greek \\ \hline
% nllb & 0.27 & 0.4 \\
% m2m100_418M & 0.59 & 0.57 \\
% EuroLLM_1_7B & 0.09 & -0.01 \\
% google_translate & 0.34 & 0.65\\
% claude & -0.03 & 0.22 \\
% \bottomrule
% \end{tabular}
% \caption{>5} 
% \label{tab:ster_g5}
% \end{table}

% \begin{table}[]
% \small
% \centering
% \begin{tabular}{c|cc}
% \toprule
% MT & French & Greek \\ \hline
% nllb & 0.97 & 0.95 \\
% m2m100_418M & 0.96 & 0.97 \\
% EuroLLM_1_7B & 0.95 & 0.91 \\
% google_translate & 0.97 & 0.99 \\
% claude & 0.95 & 0.98 \\
% \bottomrule
% \end{tabular}
% \caption{<3} 
% \label{tab:ster_s3}
% \end{table}

% \begin{table}[]
% \small
% \centering
% \begin{tabular}{c|cc}
% \toprule
% MT & French & Greek \\ \hline
% nllb & 0.94 & 0.93 \\ 
% m2m100_418M & 0.97 & 0.94 \\ 
% EuroLLM_1_7B & 0.9 & 0.78 \\ 
% google_translate & 0.96 & 0.99 \\ 
% claude & 0.89 & 0.96 \\ 
% \bottomrule
% \end{tabular}
% \caption{3=><=5} 
% \label{tab:ster_se5_ge3}
% \end{table}

\subsubsection{Divergence from the Real World}
Our findings show that MT systems do not simply reflect real-world gender imbalances—they often amplify or even distort them. While gender gaps in certain occupations still exist, the models tend to exaggerate these differences or, in some cases, completely reverse them. For instance, most systems translated texts related to `Administrative and Specialised Secretaries' (ISCO code 334) predominantly into masculine forms in French—over 80\% of the time—despite the fact that in 2023, more than 90\% of people in this occupation in France were women. Additionally, as shown in Table~\ref{tab:mt_performance} most systems produce masculine forms nearly twice as often as what real-world statistics suggest. 

While true gender equality in the labor market is still far from reality, recent data shows clear progress in reducing gender segregation across occupations. Women today participate in a much broader range of professions than in the past, and the overall numbers of employed men and women are approaching balance in many countries. However, the behavior of MT systems does not reflect this progress. Instead, their outputs often resemble labor patterns from decades ago, when women were largely confined to a limited set of roles such as nurses, or cleaners. This means that even if the models themselves are not getting worse, they diverge more and more over time from the real world because society moves forward, while the systems remain stuck in outdated patterns. As a result, the gap between model outputs and present-day labor realities slowly grows.

To better understand what shapes model behavior, we compared how closely the model outputs align with gender stereotypes versus real-world labor statistics (see Appendix~\ref{app:correlation_real_stereotype}). We found that the correlation with stereotypical perceptions is slightly—but consistently—higher than with actual employment data. 
Stereotypes often reflect outdated or oversimplified views of gender roles, and their influence on model behavior points to deeper biases in the underlying datasets. As widely acknowledged in the literature \cite{leavy2020mitigating, bender2021dangers}, training data frequently underrepresents female, minority, and non-Western perspectives, while favoring sources that reinforce dominant norms. These imbalances in representation—and in how information is structured—can amplify stereotypical associations. Importantly, even if training data were to perfectly mirror present-day labor statistics, models might still form overly rigid associations, such as consistently linking certain jobs with one gender. This highlights that data alone is insufficient to prevent biased behavior; model architecture, training objectives, and design decisions also play a crucial role.

\subsubsection{Bias Alignment}
To examine whether gender biases in MT systems are shared across languages, we computed the correlation of gendered translation distributions between the two target languages for each model. Most models showed strong cross-lingual correlations (mean $r = 0.757 \pm 0.140$), indicating that their gender biases are largely consistent across languages. This may suggest that many systems may rely on a shared internal representation that transfers similar gender preferences across languages, or simply that language and people share common gender biases across languages. NLLB-1.3B exhibited a notably lower correlation ($r = 0.478$), which aligns with its overall lower bias and reduced preference for masculine defaults as indicated in Table~\ref{tab:mt_performance}. This may indicate a more language-specific approach to gender, rather than a shared cross-lingual bias. Additionally, NLLB-1.3B showed consistently lower alignment with other models across individual languages, while the remaining models were more similar to each other. Full correlation scores are provided in Appendix~\ref{app:correlations}.

\section{Conclusions}
In this work, we explored the evaluation of MT systems when translating gender-ambiguous occupational terms. We introduced a pipeline to detect gender assignments as an indicator of potential gender bias and proposed a probability-based metric to quantify this bias against reference distributions. This approach allows for evaluation against normative standards, such as equal gender representation, as well as real-world distributions. Additionally, we provided a comprehensive benchmarking dataset containing nearly 10,000 English texts with gender-ambiguous occupational terms, covering the entire ISCO-08 spectrum of occupations. Using this framework, we evaluated 6 widely used MT systems with diverse characteristics, demonstrating the valuable insights our approach can provide.

Future work will focus on adapting the methodology to un-annotated texts, enabling gender bias evaluation of MT datasets and expanding the analysis to more languages. Furthermore, we aim to expand our framework to be able to evaluate gender-neutral translations  as well, aligning with current efforts in the field to promote inclusivity and responsible language generation.

\section*{Acknowledgments}
We want to thank Markella Challiori for introducing us to the debate on normative correctness versus predictive accuracy, which informed the framing of this work, and for her valuable insights during our discussions on gender biases in AI models.

This work was carried out within the framework of the Pharos AI Factory project, funded by the European High-Performance Computing Joint Undertaking (EuroHPC JU) under Grant Agreement No. 101234269 as part of the Horizon Europe and by the Greek Public Investments Program programme.

This work was supported by the FCT project ``OptiGov'', ref. 2024.07385.IACDC (DOI 10.54499/2024.07385.IACDC), funded by the PRR under the measure RE-C05-i08.m04.

\section*{Limitations}
A limitation of our work is the use of AI systems that may themselves be biased to detect gender biases in MT. However, we employ these systems for more narrowly defined tasks—namely, occupation detection and gender attribution—that are comparatively simpler and less ambiguous than the overall MT task being evaluated. This focused application reduces the likelihood of the models' inherent biases significantly impacting our results, as also evidenced by our method's near-perfect accuracy in occupation and gender detection.  Additionally, it is worth noting that many state-of-the-art MT evaluation metrics, such as COMET \cite{rei-etal-2020-comet}, are themselves based on large language models, which further supports the suitability of LLMs for evaluating translation quality and related properties. This alignment with established practices underscores the reliability of using LLMs in our evaluation framework.

Furthermore, a limitation of our work is the use of an artificially created dataset, which, while properly curated and manually inspected, may still carry some inherent constraints. We acknowledge that relying on such a dataset could introduce biases or limitations in terms of its representation of real-world data. However, this was the only viable option, as no existing dataset with the necessary characteristics for our study—specifically one that includes gender-ambiguous occupational terms across a wide range of occupations—was available. Despite this, the careful curation and expert review of the dataset aimed to minimize potential issues and ensure its reliability for the purpose of our analysis.

Another limitation is our treatment of gender as a binary feature, despite the growing recognition of gender as a spectrum as well as technical approaches to gender-neutral translations. From a grammatical perspective, our classification into masculine, feminine, and ``not clear'' partially addresses this complexity to some extent. However, this binary approach remains an oversimplification that fails to capture the full diversity of gender identities, which could be potentially harmful. Nonetheless, real-world statistics are predominantly published using a binary gender framework, making it challenging to analyze this issue in a more nuanced way.

Lastly, as societal roles evolve, so do occupations. Emerging professions, such as content creator or prompt engineer, may not be adequately represented in the ISCO-08 classification and thus are not fully captured in our analysis. In future work, we aim to incorporate these ``emerging occupations'', as discussed in the literature, to provide a more comprehensive evaluation of gender biases across the occupational spectrum.

\section*{Ethical Considerations}
In conducting this work, we acknowledge the ethical responsibility of ensuring our methods accurately detect gender bias to avoid unintentionally contributing to ``fairwashing''—the portrayal of biased models as fair. To address this, we intentionally simplified our approach and metrics to maintain transparency in our methodology. This design choice ensures that, even if certain components of our pipeline do not perform as expected, the rationale behind each step remains clear, facilitating easy investigation of any irregularities that could compromise the integrity of our approach. Furthermore, we evaluated our method across a broad range of occupations, recognizing the importance of capturing diverse contexts to provide a more comprehensive and ethically sound analysis of gender bias in machine translation systems.

% \begin{table}[]
% \small
% \begin{tabular}{l|ccc}
% \toprule
%         & Level 2 & Level 3 & Level 4 \\ \midrule
% English &  &  &     \\
% French  & &    &   \\
% Greek   &  &    &  \\ \bottomrule
% \end{tabular}
% \end{table}

% \begin{table}[]
% \small
% \begin{tabular}{c|ccc}
% \toprule
%         & Occ Acc & Isco Matching Acc. & Gender Acc. \\ \midrule
% English & 82.75 & 64.07     & 77.96       \\
% French  & & 52.94     & 58.82       \\
% Greek   & & 71.79     & 66.66      \\ \bottomrule
% \end{tabular}
% \end{table}

\bibliography{custom}

\appendix

\section{Implementation details}
\label{app:tech-details}
For the generation of the pipeline validation dataset, the benchmarking dataset and the pipeline for extracting occupations along with their genders, we utilized Claude-Sonnet-3.5 v2\footnote{\url{https://openrouter.ai/anthropic/claude-2}}. The MT systems evaluated in this work are presented in Table \ref{tab:model-card}. For detecting occupations, we applied a cosine similarity threshold of 0.8 when comparing the LLM-generated descriptions with ISCO-08 entries. As described in the main text under our occupation detection procedure, any detected term with a similarity score below this threshold was discarded as a hallucination.

\begin{table*}[h!]
\centering 
\small
\begin{tabular}{l|c}
\toprule
MT name & URL \\ \midrule
NLLB 600M& \url{https://huggingface.co/facebook/nllb-200-distilled-600M} \\
NLLB 1.3B& \url{https://huggingface.co/facebook/nllb-200-1.3B} \\
M2M100 & \url{https://huggingface.co/facebook/m2m100_418M} \\
EuroLLM & \url{https://huggingface.co/utter-project/EuroLLM-1.7B} \\
GT & \url{https://pypi.org/project/googletrans/} \\
Claude & \url{https://openrouter.ai/anthropic/claude-2} \\
\bottomrule
\end{tabular}
\caption{Machine translation (MT) models used, along with their corresponding hyperlinks.}
\label{tab:model-card}
\end{table*}

\section{Prompts}
\label{app:prompts}

The prompt used by our generation process is presented below.

% \par\noindent\rule{0.5\textwidth}{0.4pt}
\small

% \small \textbf{prompt} = ``````
\begin{tcolorbox}[colback=gray!5!white, colframe=black!75!black, title=Generation prompt, fonttitle=\bfseries, sharp corners=south]
Generate a <category> that explicitly mentions the occupation '<occupation title>' in its correct context. Keep it concise. Ensure that no other occupations are mentioned in the text. Ensure the occupation is referred to in a <gender> way, using pronouns, direct mentions, or other linguistic cues. 
\end{tcolorbox}

% \par\noindent\rule{0.5\textwidth}{0.4pt}
\normalsize 
The category refers to one of the text types, namely short stories, brief news reports, short statements, short conversations, and short presentations. The occupation title is provided exactly as listed in ISCO-08. For example, an occupation is ``City Councillor.'' For the benchmarking dataset (GAMBIT), the \texttt{<gender>} parameter was always set to \emph{Not Clear}, while for the evaluation dataset it was set to either \emph{Masculine} or \emph{Feminine}. Lastly, for text generation in different languages (Greek and French) for the evaluation dataset, the phrase ``The text should be in <language>.'' is appended at the end, where <language> is either ``Greek'' or ``French.''

For the extraction of the occupation and gender identification, two separate messages were provided in the same chat. The first message identifies the occupation and provides a description that can be used for matching with ISCO’s description, while the second message is for gender identification. These messages are presented below. The examples used for few-shot learning were fixed and always provided in the corresponding language (Greek or French), but here we present the prompt with the examples translated into English for clarity.

% \par\noindent\rule{0.5\textwidth}{0.4pt}
\small

\begin{tcolorbox}[colback=gray!5!white, colframe=black!75!black, title=Message 1, fonttitle=\bfseries, sharp corners=south]
In the following text, identify the occupation titles that are explicitly stated and provide the occupation title along with a brief definition in the following format: \\
\\
Occupation title: [Occupation title exactly as it is referred to in the text]\\
Definition: [Definition] \\
If no occupation is identified, please respond with: \\
"No occupation found." \\
\\
Here is an example: \\
\\
Text: \\
He is a butcher and he is a lawyer. \\
Occupation title: Butcher \\
Definition: <definition> \\
Occupation title: Lawyer \\
Definition: <definition> \\
\\
Text: \\
<text>
\end{tcolorbox}

\begin{tcolorbox}[colback=gray!5!white, colframe=black!75!black, title=Message 2, fonttitle=\bfseries, sharp corners=south]
Please now provide the gender of each identified occupation.\\
Select from one of the following options:\\
\\
Masculine if you identified in the text that the occupation refers to a masculine gender.\\
Feminine if you identified that the occupation refers to a feminine gender.\\
Not clear if, based on the text, you cannot determine the gender of the occupation.\\
You must be certain before providing the gender of the occupation and have a clear indication of its gender. \\
\\
You must answer using only one of the three options and nothing else.\\
\\
For example: \\
 \\
Text: He is a butcher, and he is a lawyer. \\
Answer: \\ 
Butcher: Masculine \\
Lawyer: Masculine \\
\\
Text: <text> \\
Answer:
\end{tcolorbox}
\normalsize

For the use of LLMs as translation systems, the prompts used for Claude 3.5 are the following:

\small
\begin{tcolorbox}[colback=gray!5!white, colframe=black!75!black, title=Translation Prompt, fonttitle=\bfseries, sharp corners=south]
Translate the following text from English to \{target\_lang\}. Provide only the translated text, without any additional context.\\
\textbf{Text:}\\
\{source\_text\}
\end{tcolorbox}
\normalsize
where target\_lang refers to the target language, either Greek or French.

For EuroLLM, we follow the template proposed in the official repository~\footnote{\url{https://huggingface.co/utter-project/EuroLLM-1.7B}}.

\section{GAMBIT}
\label{app:gambit}
Table \ref{tab:length_character_en}\footnote{Tokenization was conducted using  \url{https://www.nltk.org/api/nltk.tokenize.word_tokenize.html}} indicates the average character and word length of each type of text contained in GAMBIT.

\begin{table}[]
\small
\centering
\begin{tabular}{l|cc}
\toprule
   Category & $Avg(|Char|)$ & $Avg(|Words|)$ \\ \midrule

Short story & 613.84 ± 78.97 & 108.13 ± 14.42 \\
Brief news report & 537.87 ± 90.35 & 85.66 ± 14.74\\
Short statement & 132.42 ± 24.96 & 20.57 ± 3.76\\
Short conversation & 326.2 ± 61.62 & 70.76 ± 12.16\\
Short presentation & 744.93 ± 143.85 & 118.43 ± 20.78\\
 \bottomrule
\end{tabular}
\caption{Average character and word length of samples per category.}
\label{tab:length_character_en}
\end{table}

\section{Pipeline Validation Dataset}
\label{app:evaluation}
The character and word length statistics for the pipeline validation datasets are shown in Table \ref{tab:length_character_fr} and Table \ref{tab:length_character_gr} for French and Greek, respectively.

In the constructed dataset, each instance comprises not only the textual content but also a set of associated metadata, including the ISCO code for the relevant occupation, its corresponding title and description (sourced from the official ISCO database), and the gender referenced within the text. This annotation enables a systematic evaluation of the pipeline’s performance by comparing the ground truth occupation and gender with the occupation and the gender predicted by the model. Specifically, for each instance we first assess the accuracy of occupation identification by verifying whether the occupations predicted by the system match the ground-truth occupation. Subsequently, we compute the accuracy of gender prediction by checking whether the gender assigned by the pipeline aligns with the ground-truth label. In cases where the system failed to detect any occupation in the sentence, the associated gender prediction was automatically considered incorrect.

\begin{table}[]
\small
\centering
\begin{tabular}{l|cc}
\toprule
   Category & $Avg(|Char|)$ & $Avg(|Words|)$ \\ \midrule

Short story & 675.15 ± 71.82 & 113.44 ± 11.99 \\
Brief news report & 546.38 ± 72.48 & 88.89 ± 11.78\\
Short statement & 147.22 ± 34.42 & 23.21 ± 5.35\\
Short conversation & 375.7 ± 61.29 & 73.75 ± 10.79\\
Short presentation & 749.73 ± 110.58 & 118.92 ± 15.93\\
 \bottomrule
\end{tabular}
\caption{Average character and word length of samples per category for the French dataset.}
\label{tab:length_character_fr}
\end{table}

\begin{table}[]
\small
\centering
\begin{tabular}{l|cc}
\toprule
   Category & $Avg(|Char|)$ & $Avg(|Words|)$ \\ \midrule

Short story & 522.56 ± 64.49 & 88.08 ± 11.2 \\
Brief news report & 457.93 ± 49.24 & 70.89 ± 8.03\\
Short statement & 127.44 ± 28.57 & 19.41 ± 4.21\\
Short conversation & 341.29 ± 67.84 & 66.06 ± 13.57\\
Short presentation & 578.84 ± 94.13 & 87.51 ± 12.82\\
 \bottomrule
\end{tabular}
\caption{Average character and word length of samples per category for the Greek dataset.}
\label{tab:length_character_gr}
\end{table}

\section{Per-model Analysis of Extreme Gender Assignments}
\label{app:extreme}

Table~\ref{tab:extreme-gendering} presents the number of ISCO occupations (4-digit level) for which each MT system exhibits either extreme or balanced gender assignments in Greek and French. We define \textit{extreme gendering} as cases where one gender is used in more than 80\% of the translations (i.e., $GRAPE^{parity}_m > 0.6$ or $GRAPE^{parity}_f > 0.6$), and \textit{balanced outputs} as those where gender assignments fall within the 30\%–70\% range. Most systems overwhelmingly favor one gender per occupation, with more than 90\% of ISCOs falling in the extreme category for several models. Google Translate and M2M100, for example, produce extreme gendering in over 420 occupations in Greek. By contrast, the large NLLB model (1.3B) shows the most balanced outputs, with over 90 occupations falling within the moderate range for both Greek and French.

\begin{table*}[h]
\small
\centering
\begin{tabular}{l|cc|cc}
\toprule
\multirow{2}{*}{MT} & \multicolumn{2}{c|}{\textbf{French}} & \multicolumn{2}{c}{\textbf{Greek}} \\ \cline{2-5}
& Extreme & Moderate & Extreme & Moderate \\
\midrule
NLLB-600M & 411 & 15 & 413 & 10 \\
NLLB-1.3B & 286 & 91 & 285 & 93 \\
M2M100-418M & 417 & 7 & 424 & 5 \\
EuroLLM-1.7B & 381 & 30 & 334 & 47 \\
GT & 321 & 6 & 427 & 5 \\
Claude & 387 & 27 & 397 & 23 \\

\bottomrule
\end{tabular}
\caption{Number of ISCO occupations showing extreme ($<$20\% or $>$80\%) or balanced (30--70\%) gender assignments across translation outputs, separated by language.}
\label{tab:extreme-gendering}
\end{table*}

\section{Stereotypes}
\label{app:stereotypes}
To quantify the gender stereotyping of occupations, we used ratings from \citet{shinar1975sexual}, which provide perceived gender associations for 129 occupations on a 1–7 scale (1 = most masculine, 7 = most feminine). Each occupation was manually mapped to its closest corresponding 4-digit ISCO category. When multiple occupations mapped to the same ISCO code, we assigned the average of their ratings to that category. If the mapped occupations exhibited substantial variability (i.e., a rating variance $\ge 1.5$), the corresponding ISCO code was excluded to ensure consistency. This process yielded 97 unique 4-digit ISCO codes, each associated with a representative rating indicating the degree of gender stereotyping.

We created the stereotypically masculine, feminine, and neutral groups for our study by grouping occupations with stereotyping rating below 2.5, above 5.5, and between 3 and 5 respectively. 

\section{Stereotypical vs. Real-World Correlations}
\label{app:correlation_real_stereotype}

To better understand the factors influencing model behavior, we examined how closely model outputs align with gender stereotypes and real-world labor statistics. Specifically, we computed the correlation between the $GRAPE^{parity}_f$ indicating the model’s predicted gender distribution and (i) the stereotype ratings described in Appendix~\ref{app:stereotypes}, and (ii) actual labor market data (female ratio). As real-world statistics are available at the 3-digit ISCO level, we aggregated both model outputs and stereotype ratings accordingly to ensure a fair comparison. For the stereotype ratings, we grouped the 4-digit ISCO codes by their first three digits and calculated the average rating for each group. To maintain consistency, we excluded any groups where the variance among constituent 4-digit occupations was $\ge1.5$. This resulted in 59 unique 3-digit ISCO groups with representative stereotype ratings. We also filtered the real-world data to retain only those 3-digit ISCO groups that appeared in the stereotype set. The detailed correlation results are presented in Table~\ref{tab:stereotype-real_corrs}. As we can see, across all models and both languages, the correlation with stereotypical ratings is consistently higher than with real-world labor statistics, suggesting a stronger alignment of model behavior with societal stereotypes than actual workforce distributions.

\begin{table}[]
\small
\centering
\begin{tabular}{c|cc|cc}
\toprule
\multirow{2}{*}{MT} & \multicolumn{2}{c|}{Real} & \multicolumn{2}{c}{Stereotype} \\ \cline{2-5}
& French & Greek & French & Greek \\ \hline
NLLB-600M & 0.47 & 0.47 & 0.5 & 0.5 \\
NLLB-1.3B & 0.41 & 0.21 & 0.41 & 0.34 \\
M2M100 & 0.31 & 0.33 & 0.34 & 0.4 \\
EuroLLM & 0.65 & 0.64 & 0.7 & 0.74 \\
GT & 0.36 & 0.29 & 0.42 & 0.39 \\
Claude & 0.63 & 0.56 & 0.69 & 0.64 \\
\bottomrule
\end{tabular}
\caption{Correlations between model outputs, real data distributions, and stereotype ratings.} 
\label{tab:stereotype-real_corrs}
\end{table}

\section{Cross-lingual and Intra-lingual Correlations}
\label{app:correlations}
To examine the consistency of gender biases across target languages, we calculated the Pearson correlation coefficient between the gender distributions produced for the two target languages (French and Greek) for each translation model. The resulting scores reflect how similarly each model assigns gendered translations across the two languages.

\begin{table}[h]
\centering
\caption{Pearson correlation between gendered translation distributions across French and Greek for each model.}
\label{tab:crosslingual_corrs}
\begin{tabular}{lr}
\toprule
\textbf{Model} & \textbf{Cross-lingual Correlation ($r$)} \\
\midrule

NLLB-600M    & 0.8563 \\
NLLB-1.3B    & 0.4775 \\
M2M100     & 0.8524 \\
EuroLLM     & 0.7482 \\
GT  & 0.7103 \\
Claude             & 0.8949 \\

\bottomrule
\end{tabular}
\end{table}

As shown in Table~\ref{tab:crosslingual_corrs}, most models exhibit strong cross-lingual correlations in their gendered translation patterns, with coefficients exceeding 0.70, suggesting largely shared gender biases across the two target languages. The only notable exception is NLLB-1.3B, whose lower correlation score ($r = 0.4775$) aligns with its generally lower gender bias and reduced reliance on masculine defaults (as discussed in Section~\ref{sec:results}). This may suggest that the model follows a more language-specific strategy for handling gender, rather than relying on shared internal representations.

\begin{figure}[h]
    \centering
    \includegraphics[width=\linewidth]{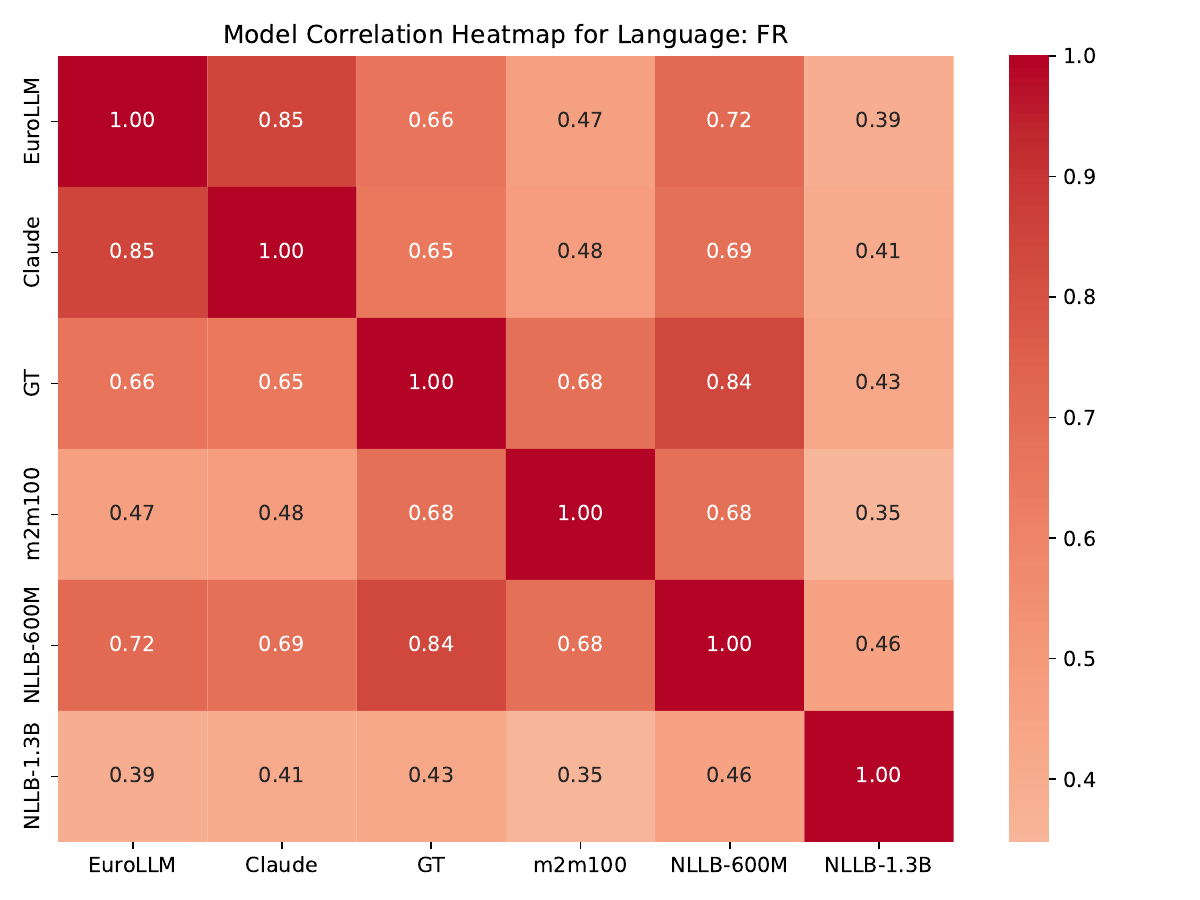}
    \caption{Intra-model correlation of gendered translation distributions in French.}
    \label{fig:intra_french}
\end{figure}

\begin{figure}[h]
    \centering
    \includegraphics[width=\linewidth]{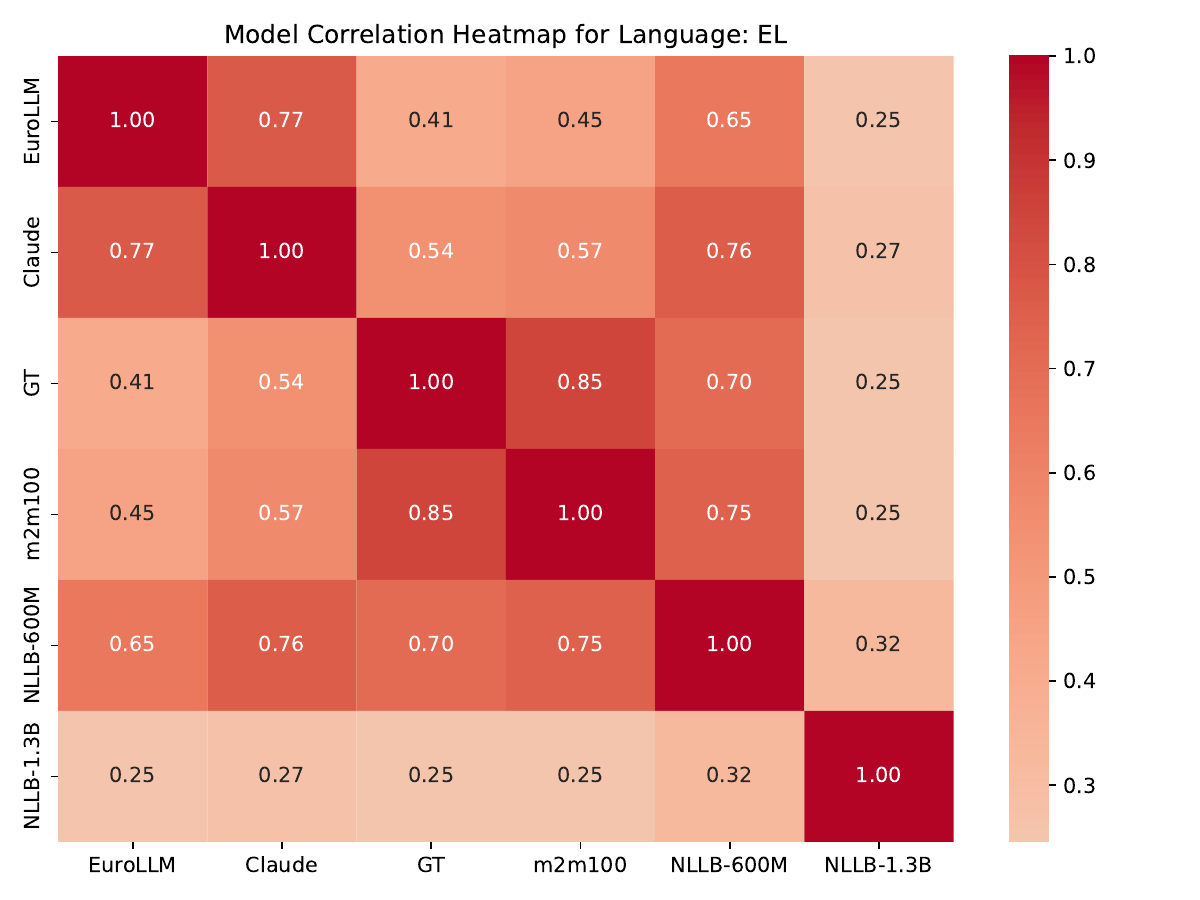}
    \caption{Intra-model correlation of gendered translation distributions in Greek.}
    \label{fig:intra_greek}
\end{figure}

To further illustrate the internal consistency of each model’s gendered behavior, Figures~\ref{fig:intra_french} and~\ref{fig:intra_greek} present intra-model correlation heatmaps across models within each language. These visualizations reveal that NLLB-1.3B also shows reduced alignment with other models in both French and Greek, reinforcing the observation that it diverges more significantly from the broader modeling trends.

\end{document}